\documentclass[11pt,a4paper]{article}
\usepackage[margin=1in]{geometry}
\usepackage{graphicx}
\usepackage{amsmath,amsfonts}
\usepackage{algorithmic}
\usepackage{algorithm}
\usepackage{array}
\usepackage[caption=false,font=normalsize,labelfont=sf,textfont=sf]{subfig}
\usepackage{textcomp}
\usepackage{stfloats}
\usepackage{url}
\usepackage{verbatim}
\usepackage{cite}
\usepackage{soul}
\usepackage{authblk}
\usepackage{placeins}
\usepackage{xcolor}

\title{Online Continual Learning in Edge-Cloud Applications}

\author[1]{Charalampos Davalas\thanks{Corresponding author. Email: \texttt{cdavalas@hua.gr}. ORCID:0000-0003-4445-4314}}
\author[1]{Dimitrios Michail\thanks{Email: \texttt{michail@hua.gr}. ORCID:0000-0002-5316-6704}}
\author[1]{Christos Diou\thanks{Email: \texttt{cdiou@hua.gr}. ORCID:0000-0002-2461-1928}}
\author[1]{Iraklis Varlamis\thanks{Email: \texttt{varlamis@hua.gr}. ORCID:0000-0002-0876-8167}}
\author[2]{Konstantinos Tserpes\thanks{Email: \texttt{tserpes@mail.ntua.gr}. ORCID:0000-0001-5183-1443}}

\affil[1]{Department of Informatics and Telematics, Harokopio University of Athens, Athens, 17778, Greece}
\affil[2]{School of Electrical and Computer Engineering, National Technical University of Athens, Athens, 15773, Greece}

\date{}

\begin{document}

\maketitle

\begin{abstract}

Deployed machine learning models in real-world IoT and edge/cloud systems face a critical reliability bottleneck: the requirement for online adaptation to non-stationary data streams while operating without access to the full original training dataset. Performance is severely degraded when models encounter Out-of-Distribution (OOD) inputs or suffer from catastrophic forgetting. Addressing these challenges requires adaptation mechanisms capable of learning from evolving data streams while preserving previously acquired knowledge.

This paper proposes an edge-cloud methodology for integrating Continual Learning (CL) with Out-of-Distribution (OOD) Detection and Concept Drift Detection in IoT applications. The methodology incorporates ODIN and ECDD as alternative mechanisms for identifying informative samples for annotation and triggering rehearsal-based model updates. Experimental results across automotive, healthcare, and agricultural datasets demonstrate the applicability of the proposed methodology and its potential to support model adaptation while mitigating catastrophic forgetting.
\end{abstract}

\section{Introduction}
The rapid growth of the Internet of Things (IoT) and edge-cloud computing has enabled numerous real-world applications—from autonomous vehicles and industrial automation to remote healthcare. 
A critical challenge in these systems is ensuring that machine learning (ML) models deployed at the edge can adapt to novel scenarios without access to the original training dataset. In practice, models often encounter out-of-distribution (OOD) inputs that differ from the data seen during initial training, leading to potential performance degradation in mission-critical applications.

A promising paradigm involves deploying adaptive models that can continually learn from local feedback. This requires a mechanism to efficiently identify novel scenarios (OOD detection), select the most informative samples for labeling (active learning), and then integrate this new knowledge without losing previously acquired skills (continual learning).
Current solutions, such as RILOD~\cite{li2019rilod} and EAR~\cite{chen2023ear}, have demonstrated feasibility. However, integrating alternative drift- and confidence-based sample-selection mechanisms into asynchronous edge-cloud continual-learning pipelines remains an important research challenge.

This paper addresses these gaps by proposing a comprehensive edge-cloud continual learning pipeline. Our design is centered on an asynchronous, rehearsal-based fine-tuning strategy, triggered specifically by robust conflict detection mechanisms. This approach prioritizes informative samples for annotation and rehearsal-based model updates, supporting continual adaptation to evolving data distributions.

The main contributions of this work are:
\begin{itemize}
    \item \textbf{A Unified Architecture for Adaptive Edge ML:} We introduce a novel, general framework that synergistically integrates OOD detection, Concept Drift detection, and rehearsal-based Continual Learning within a practical edge-cloud deployment pipeline.
    
    \item \textbf{Drift and Out-of-Distribution Sampling Strategy:} We evaluate model-versus-user disagreement and model confidence scoring as alternative mechanisms for selecting informative samples and supporting continual model adaptation in IoT environments.

    \item \textbf{Performance and Robustness Analysis:} We evaluate the proposed methodology across automotive, healthcare, and agricultural applications, examining its ability to adapt to new stream data while retaining previously acquired knowledge.
\end{itemize}

The remainder of this paper is organized as follows. Section II reviews related work on online continual learning, edge-cloud adaptation, and OOD detection. Section III presents the proposed framework, including continual learning strategies and adaptive model updates. Section IV describes the experimental evaluation, while Sections V and VI illustrate applications in automotive and healthcare domains, respectively. Section VII concludes with discussion and insights.

\section{Related Work}\label{sec:relwork}

The integration of continual learning methodologies into Internet of Things (IoT) applications, particularly within edge–cloud architectures, has recently received significant attention. This research direction addresses the challenges of real-time data streams, limited access to original training datasets, and strict operational constraints.

\subsection{Edge Computing in IoT}
The Once-for-All (OFA) network \cite{cai2019once} proposes a strategy for efficient deployment of deep neural networks across diverse hardware platforms. OFA effectively addresses model compression and deployment heterogeneity.
Most traditional edge computing frameworks emphasize computational efficiency in resource-constrained environments but often lack mechanisms for dynamic model updating. Zhang et al.~\cite{Zhang2019light} employed a dual-network setup for traffic sign classification at the edge, utilizing knowledge distillation to balance supervisory and decision-making tasks. Similarly, Tabernik et al.~\cite{Tabernik2020sign} developed efficient traffic sign detection pipelines, while Ali et al.~\cite{Ali2020iot} extended these concepts to large-scale IoT deployments across urban environments.  
More recently, frameworks such as RILOD~\cite{li2019rilod}, which incrementally constructs datasets ``on-the-fly'' to learn new classes, and EAR~\cite{chen2023ear}, which adapts models via shallow adaptors to mitigate catastrophic forgetting under resource constraints, have demonstrated the feasibility of embedding continual learning into edge--cloud systems. These works highlight the need for scalable architectures that support not only real-time inference but also efficient model adaptation in dynamic environments.  The work of \cite{zeng2025online} focuses of scheduling pretrained models based on directed acyclic graphs and edge availability. The models are not necessarily trained continually. Finally AdaInf \cite{shubha2023adainf} focuses on similar principles for scheduling training and inference of models based on model drift.

\subsection{Continual Learning and Catastrophic Forgetting}
A long-standing challenge in adapting continual learning to IoT environments is catastrophic forgetting, where models lose previously acquired knowledge when adapting to new information. Traditional continual learning methods, such as those proposed by Rebuffi et al.~\cite{rebuffi2017icarl}, Hoiem et al.~\cite{Hoiem2016lwf}, and Kirkpatrick et al.~\cite{kirkpatric2017ewc}, primarily address scenarios with predefined task boundaries. However, in real-world applications, especially in IoT contexts, environmental changes are often unpredictable, necessitating task-agnostic learning approaches.

Chaudhry et al.~\cite{chaudhry2019continual} introduced a combination of experience replay with reservoir sampling, significantly outperforming prior techniques in adaptive, task-agnostic learning pipelines. Prabhu et al.~\cite{prabhu2020gdumb} revisited rehearsal methods, demonstrating competitive performance with a simplified and adaptable framework.

Our work extends these methodologies by emphasizing the integration of continual learning techniques for real-time model adaptation in streaming data, focusing on detecting and responding to distributional variations without relying on predefined task boundaries.

\subsection{Online Continual Learning}
Online continual learning extends these ideas by emphasizing real-time adaptation. Recent methods, such as the work of Pellegrini et al.~\cite{pellegrini2020latent}, emphasize the real-time aspect of learning. These approaches preserve and reuse a latent space of activations for rehearsal training, avoiding the need to store raw data. When combined with existing continual learning techniques, this strategy has shown promising results in experiments conducted on datasets like OpenLoris and CORe50~\cite{shi2020we,lomonaco2017core50}.

In our previous work~\cite{davalas2024rehearsal,davalas2022computationally}, we proposed a rehearsal-based method to mitigate catastrophic forgetting in a resource-constrained online setting. This method combines a drift detection mechanism with a rehearsal strategy similar to experience replay. A key distinction is that, instead of retraining continuously, the model is updated only when the drift detector raises a flag. Furthermore, we introduced an adaptive rehearsal strategy that dynamically determines the number of training steps based on the detected drift magnitude.

Building on this foundation, we extend these methods and integrate them into an edge--cloud architecture for online learning, enabling dynamic model updates in IoT applications.

\subsection{Drift Detection Mechanisms}
Drift detection methods are central to adaptive streaming systems, as they identify gradual or sudden distributional changes. Influential methods in this field include ADWIN~\cite{bifet2007adwin}, which uses windowing in a data stream based on drift per time step, and ECDD~\cite{ross2012exponentially}, which combines exponentially weighted moving average tables with statistical drift detection, offering computational efficiency.

While drift detection is potent, as shown in our previous work~\cite{davalas2022computationally,davalas2024rehearsal}, it can be further improved by utilizing methods that provide model diagnostics without requiring ground truth labels. This enhancement is crucial for real-time applications where obtaining labeled data is challenging.

\subsection{Out-of-Distribution Detection}
Out-of-distribution (OoD) detection refers to a model's ability to identify input samples that differ from those seen during training and handle them appropriately. OoD methods are particularly valuable in safety-critical applications such as autonomous driving, where reliable model confidence is essential~\cite{seppanen2024out}.

OoD methods can be broadly categorized into classification-based, density-based, reconstruction-based, and distance-based approaches. Among these, the ODIN method~\cite{liang2018enhancing} perturbs inputs using gradient-based noise and evaluates the resulting softmax confidence to detect distributional shifts, without requiring modifications to a pre-trained model or ground truth labels. Considering the use of OOD methods in IoT Systems, RILOD~\cite{li2019rilod} employs selective labeling of OOD samples to incrementally learn new classes, while EAR~\cite{chen2023ear} introduces lightweight adaptors to address domain shifts. These systems illustrate how OOD detection can act as a trigger for continual fine-tuning, ensuring both adaptability and robustness in dynamic IoT environments.

In our work, we adapt OoD detection for online continual learning to mitigate catastrophic forgetting by training on samples exhibiting distribution shifts, rather than relying on arbitrary hyperparameters or randomized sampling. 

\section{Our Approach: An edge-cloud online continual \\learning pipeline}\label{sec:proposition}

Smart applications have been developed to support various tasks such as autonomous driving, dashcam monitoring, disease detection, and quality assurance. Despite their utility, the models and devices employed in these systems often lack the ability to adapt to novel conditions or environmental changes.

To address this limitation, we propose the incorporation of an additional edge-cloud interaction layer that enables continual learning. By leveraging modern rehearsal-based techniques, the neural network model can be incrementally updated without requiring large volumes of training data—unlike conventional full retraining pipelines. An overview of the proposed framework is presented in Fig.~\ref{fig:base}.

{
\begin{figure*}[htbp] \centering \includegraphics[width=0.8\textwidth]{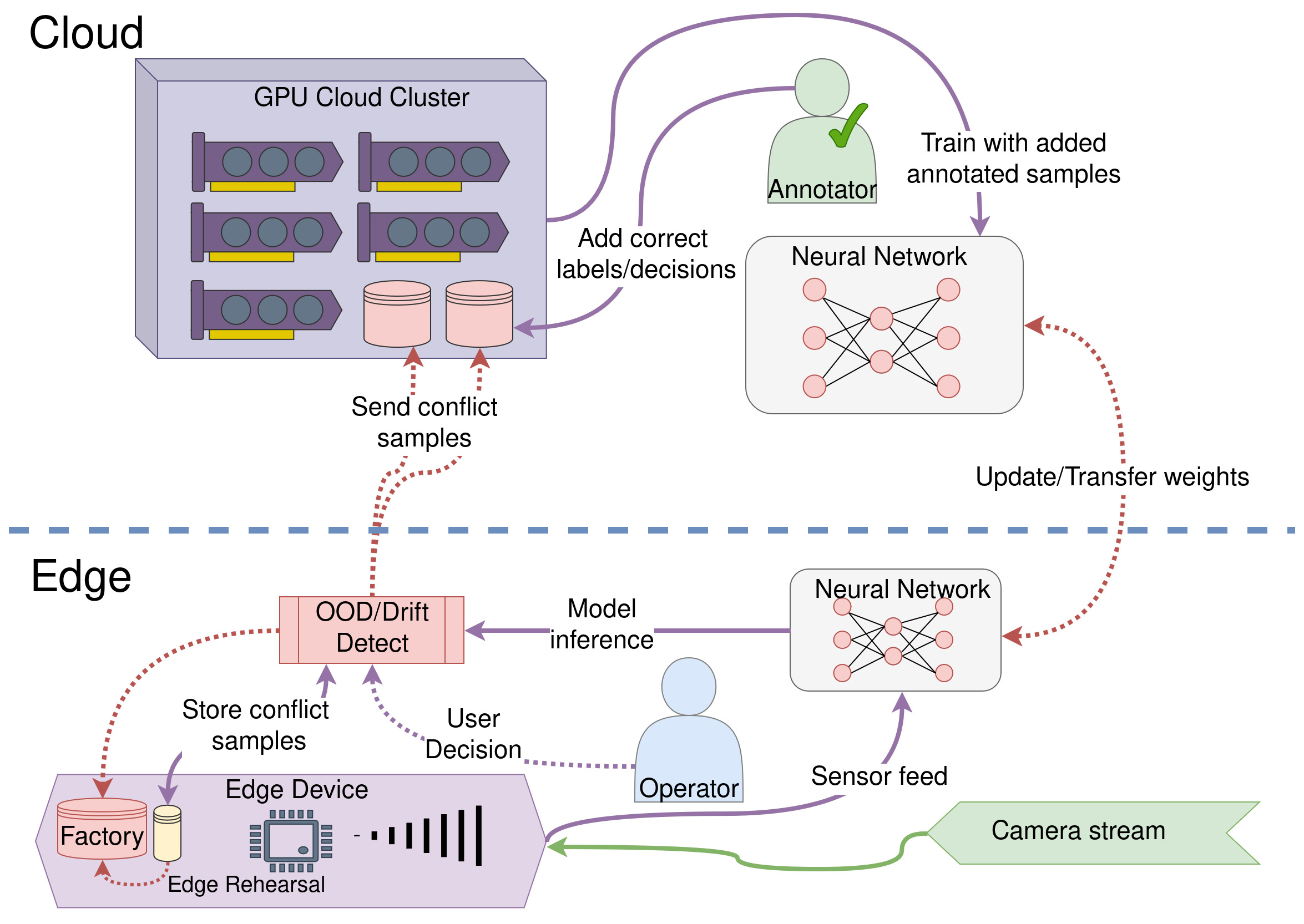} \caption{Overview of the proposed continual learning pipeline using edge-cloud interactions. The edge device includes a mechanism for identifying novel or ambiguous samples, which are stored for deferred processing in the cloud. Labels are provided or verified by a human annotator, and the model is updated through a continual learning mechanism before being redeployed to the edge.} \label{fig:base} \end{figure*}
}

{
This setup implements a supervised, continual learning pipeline, with processing distributed between the edge and the cloud infrastructure. Initially, both the operator and the model observe the environment and make decisions based on the sensor input. We assume that the sensor data is noise-free and that the observation space is unobstructed.
}

{
The operator's actions are expressed as labels, representing fundamental decisions—such as decelerating or stopping a vehicle, or providing a positive or negative diagnosis in a medical setting. The model generates suggestions in the same label space.
}

{
Occasionally, discrepancies arise between the operator’s decisions and the model’s predictions. These can result from human error, inattention, or from model inaccuracies due to unfamiliar or low-visibility inputs. Upon repeated conflict, a drift or out-of-distribution (OoD) detector is triggered, and the corresponding samples are stored in a buffer without labels.
}

{
Subsequently, the buffered samples are transmitted to a cloud infrastructure equipped with the required computational, memory, and human resources. A domain expert reviews and annotates the samples, providing unbiased ground truth labels.
}
{
The newly annotated data is then used to retrain the model using a GPU-enabled cloud infrastructure. The updated model is subsequently deployed back to the edge. Details of the edge-side and cloud-side operations are discussed in Sections~\ref{edge} and~\ref{cloud}, respectively.
}
\subsection{User and Model Conflicts detection}\label{edge}

Each edge device stored data based on a specific conflict detection algorithm in order to keep a sufficient amount of possibly questionable data in a buffer. The methods of data storage can be either based on drift detection between driver input and model output or by Out-of-Distribution detection. By using this mechanism, the benefits are twofold. First, model updates are initiated according to detector signals and buffer conditions rather than being performed continuously, extending the event-driven adaptation strategy of our previous work in \cite{davalas2024rehearsal}. Secondly, the model is only trained in specific intervals and in the case of drift/OoD detection, the model will be trained based on its classification statistics and not only based on the statistical properties of the data. In the following subsections, we give a detailed description of the anomaly detectors in use.

\subsubsection{Drift detection}

One of the most well established methods for concept drift, is the use of the Exponentially Weighted Moving Average Charts for Detecting Concept Drift (ECDD) \cite{ross2012exponentially}. This method combines the Exponential Weighted Moving Average (EWMA) Charts with a change detector based on the Bernoulli distribution. This method requires two streams for detecting drifts between them, and a flag is raised when that happens. 
the triggering factor for raising the alarm is based on three conditions i) a drift is detected, and the buffer is non-empty, ii) the dynamic buffer is filled and iii) while having a non-empty buffer, a certain number of time steps t has passed without any drift, therefore the model was for a significant amount of time idle in terms of training. The use of ECDD, can provide a stable and computationally effective alternative for continual learning with limited resources ($O(1)$ complexity). One issue that is possibly detrimental to the use of such a method is the fact that ECDD requires both driver reactions and model output. We also provide an algorithmic description of ECDD for clarity in algorithm \ref{alg:ECDD} based on \cite{ross2012exponentially}

\begin{algorithm}
\caption{\texttt{ECDD}}
\label{alg:ECDD}

\textbf{Input:} Model logits $y_i$, ground truths $y$, decay factor $\lambda$, average run length $ARL$, $L_t$ precomputed statistics tables\\
\textbf{Output:} Drift boolean flag $drift\_warn$

\vspace{1em}

1.\quad Initialize: $drift\_warn \leftarrow \text{False}$, $Z_0 \leftarrow 0$, $\hat{p}_{0,0} \leftarrow 0$\\
2.\quad \textbf{For each} $(y_i, y)$ \textbf{do:}\\
3.\qquad $X_t \leftarrow 
\begin{cases}
0, & \text{if } \arg\max(y_i) = \arg\max(y) \\
1, & \text{otherwise}
\end{cases}$\\
4.\qquad $\hat{p}_{0,t} \leftarrow \dfrac{t}{t+1} \hat{p}_{0,t-1} + \dfrac{1}{t+1} X_t$\\
5.\qquad Compute $\hat{\sigma}_{X_t}$ and $\hat{\sigma}_{Z_t}$ from $\hat{p}_{0,t}$\\
6.\qquad $Z_t \leftarrow (1 - \lambda) Z_{t-1} + \lambda X_t$\\
7.\qquad \textbf{If } $Z_t > \hat{p}_{0,t} + L_t \hat{\sigma}_{Z_t}$ \textbf{ then:}\\
8.\qquad\qquad $drift\_warn \leftarrow \text{True}$\\
9.\quad \textbf{Return } $drift\_warn$

\end{algorithm}

\subsubsection{Out of Distribution detection}
Recent advancements in anomaly detection have focused on monitoring model status and determining when retraining is required, particularly when incoming samples deviate from the training data distribution. Out-of-Distribution (OoD) detection methods offer the advantage of evaluating a model’s decision-making capability without requiring external input. In driving applications, for example, the model can report its own status solely based on sensor data, eliminating the need for driver feedback.

One widely used algorithm for this purpose is ODIN \cite{liang2018enhancing}, which provides a metric to assess a model’s classification confidence. ODIN is model-agnostic and consists of two primary components: an input preprocessing routine that perturbs incoming samples, and a temperature-scaled softmax function that evaluates the model’s response. Together, these produce a confidence score, which is then compared against a threshold to determine whether a sample is considered low-confidence.

\noindent \textbf{Out-of-Distribution Detection via ODIN:} 
ODIN is a post-processing technique that enhances the separability between samples using temperature scaling and small input perturbations. Specifically, given a neural network producing logits \( f(x) \), ODIN rescales them by a temperature parameter \( T > 1 \) before
a small, targeted perturbation is added to the input. The perturbed and temperature-scaled confidence is used as the OOD score: higher values indicate in-distribution samples. ODIN provides an OOD score without requiring the underlying model to be retrained for detection.

Unlike ECDD, ODIN operates independently of user input, making it more adaptive to dynamic environments. In Equations \ref{eq:softmax_t}, \ref{eq:pertube_x}, and \ref{eq:ood}, we provide a formal definition of the ODIN algorithm.

\begin{equation}\label{eq:softmax_t}
	S_i(x;T) =\frac{exp(f_i(x)/T)}{\sum_{j=1}^{N}exp(f_j(x)/T)}
\end{equation}
\begin{equation}\label{eq:pertube_x}
	\hat{x} = x-\epsilon sign(-\nabla_x log(max_{i}(S_i(x;T))))
\end{equation}
\begin{equation}\label{eq:ood}
	OoD(x; \delta, T, \epsilon) = \begin{cases}
		1,&  max_{i}(S_i(\hat{x};T)) \leq \delta \\
		0,&  max_{i}(S_i(\hat{x};T)) > \delta    \\
	\end{cases}
\end{equation}

In table \ref{tab:alg_comparison} we provide the pros and cons of each algorithm. Note that directly comparing ECDD and ODIN is not applicable due to the difference in their original roles.

The \(O(g)\) is the complexity of calculating the model gradients and \(O(f)\) is the model inference which both methods require in order to evaluate the model logits. Note that these complexities are highly affected by the model architecture. 

ODIN can be less scalable in larger architectures due to the additional overhead of input perturbations and temperature scaling. However, it is generally safe to assume that IoT applications often employ smaller or lightweight models, making this overhead more manageable in such contexts.

\begin{table}[hbtp]
\caption{ECDD and ODIN comparison.}
\label{tab:alg_comparison}
\centering
\begin{tabular}{ccc}
    Algorithm & Complexity & Requires ground truth?\\
    \hline
    ECDD & \textcolor{green}{\(O(f)+O(1)\)} & \textcolor{red}{Yes}\\
    ODIN & \textcolor{red}{\(O(f)+O(g)\)} & \textcolor{green}{No}\\
\end{tabular}

\end{table}
\subsection{Online continual learning on the cloud}\label{cloud}
An edge device is constrained in terms of hardware. Even the most lightweight neural networks require a significant amount of computation and memory, which may not be readily available \cite{liu2021efficient}. Methods such as reducing parameter precision (e.g., from 64-bit to 32-bit floating-point) can lower computational overhead; however, a model will eventually require updates to accommodate new data. An online continual learning pipeline can support model updates in response to changes in incoming data.

Our approach utilizes a rehearsal method with conflict-based sampling. Instead of selecting samples based on occurrence (e.g., reservoir sampling), we employ a selection process that detects and prioritizes conflict-inducing samples. These may be identified either by triggering a drift detection mechanism or through out-of-distribution (OoD) detection.

In parallel with sample selection, the selected detection mechanism determines when the buffered samples should initiate a model update. Drift detection uses disagreement between model predictions and user responses, whereas OOD detection identifies samples that differ from the model's training distribution. In our setup, these alternative mechanisms guide sample selection and the timing of rehearsal-based updates.

In the case of drift detection, each batch of samples is passed through the neural network. The model’s outputs are compared against user responses, and disagreements are aggregated into a classification score. If the model and user repeatedly choose different actions—indicating a potential drift—an alarm is raised. These drift-inducing samples are retained, and the detector's statistics are used to trigger further actions when stream samples deviate significantly from expected user responses. For OoD detection, a buffer limit is used: if the buffer is filled with samples flagged as out-of-distribution, this serves as a valid signal without requiring user input.

A key advantage of this method is its flexibility: if hardware permits, the setup can be deployed on either edge or cloud infrastructure interchangeably. By combining selective sample acquisition with rehearsal-based learning, the proposed pipeline supports model adaptation using a curated subset of incoming stream data. Another benefit is the potential for each edge device to contribute useful data, such as parts of a camera feed or locally trained model snapshots. These can later be aggregated or annotated—either through human input or more advanced automated mechanisms—addressing the challenge of ground truth scarcity in sensor-driven applications.

\section{Experimental setup}\label{sec:expset}
In this section, we show the outline of our propositions as shown in Fig \ref{fig:base}. The basic concept is to propose and test methods both within an edge device and on a cloud infrastructure. We analyze the setup of experimentation for a variety of Cloud-Edge interactions for continual learning from sensor input. The task is to test classification accuracy of models with our proposed continual learning framework. The full experimental setup and the implementations have been developed with TensorFlow2 \cite{tensorflow2015-whitepaper}

\subsection{Method combinations}\label{methods_section}
Testing all the possible method combinations we propose combinations based on the methodologies used for conflict resolution alongside various methodologies for rehearsing new samples along with older ones. In Table \ref{methods_list}, we refer to those combinations in detail.
\begin{center}
\renewcommand{\arraystretch}{0.5} 
\begin{table*}
\caption{List of all methodologies with abbreviations and a short description. Note that, in the case of edge training there is no sample selection. The methods in asterisk require user input to the edge device/model}
\label{methods_list}
\setlength{\tabcolsep}{3pt}
\renewcommand{\arraystretch}{1.05}
\small
\begin{tabular}{ | p{0.29\linewidth} | c | p{0.6\linewidth} | }

\hline
Training methodology & Abbr. & Description  \\
\hline
 Edge training* & TRN & The model is being trained only with new samples and the user/annotator labeling. This type of continual learning usually induces catastrophic forgetting.\\ 
\hline
 Edge/cloud rehearsal & RH & The model is being trained with rehearsing new samples (with their respective labels/annotations), alongside older samples from a factory dataset subset (memory buffer). This operation can either reside in an edge device or a cloud infrastructure\\ 
\hline
 Cloud annotation and dataset expansion & ANDE & The model is being trained with the newly annotated samples and an extended dataset which is composed of factory samples and older annotated images from another edge device   \\ 
   \hline

   Sample selection method & Abbr. & Description \\
\hline
 Random & RAND &The samples are being loaded in a buffer with reservoir sampling regardless of conflict or not.  \\
 \hline
 Drift detection* & ECDD & Drift detection method as described in \ref{edge}. The samples are being stored, based on the disagreements between the operator and the model. The samples are being used for rehearsal when the method is triggered or when the buffer is full. This method requires user input\\ 
 \hline
 Out-Of-Distribution detection & ODIN & Out-Of-Distribution method as described in \ref{edge} the OoD samples are stored in an edge buffer and are being used when the buffer has reached its limit.\\ 
  \hline 

\end{tabular}
\end{table*}
\end{center}
\subsection{Factory and Stream accuracy}

In order to test if a model's ability to maintain current knowledge
we need a simple accuracy metric based on held-out samples, based on the knowledge of the model from the pre-training phase, named \textbf{Factory accuracy}.
For testing a model's ability to expand knowledge, we use a second metric is actually an accuracy metric based on a combination of samples which are also held out and are originated from the stream samples. We name this metric \textbf{Stream accuracy}. Both metrics do not differ from the original classification accuracy A which is 
\[ A = \frac{\#Correct\ predictions }{\#Total\ predictions}  \]
The only difference is the data we use in order to calculate those metrics.

We have run the tests three times per method for reproducibility, each method has a mean-standard deviation for each time step, which will be shown in the following figures (\ref{fig:belgium_edge}, \ref{fig:nuscenes_edge}, \ref{fig:malaria_edge}, \ref{fig:plant_village_edge}). The scores are presented as the mean-standard deviation of the total average of scores per testing run which is provided by the following tables \ref{tab:methods_acc_belgium}, \ref{tab:nuscenes_edge}, \ref{tab:malaria_edge} and \ref{tab:plant_village_edge}

\subsection{Models in use}
Excluding the ``toy scenario'' experiment (section \ref{belgium}), the base model in our experiments is the MobileNet, used in the majority of the edge-cloud learning literature \cite{howard2017mobilenets}. We aim to classify smaller image samples under the assumption that the camera is not necessarily the best in terms of analysis. The model is first trained offline with a subset of the dataset. The model used in the experiments is a ``factory pre-trained'' model which is trained, based on a preset amount of samples that cannot include all possible variations of real life samples. This model is used at the start of online training in all experiments.

\section{Assisted driving}\label{sec:car}
Although the 1968 Vienna Convention on Road Signs and Signals \footnote{https://unece.org/DAM/trans/conventn/signalse.pdf} was the first attempt to agree on a universally accepted format of traffic signals, and a lot of progress has been performed since then, there are still differences among countries, in addition to maintaining local peculiarities, which makes the task of traffic sign detection systems even harder. 

A smart detection system must be able to respond to such a dynamic environment and be able to perform under a variety of cases. This is usually achieved by augmenting the training datasets with variations of the training samples or expanding them with more samples from other datasets, thus widening the problem space coverage. Although such actions can improve the system performance, they can become very complex and cost-inducing in the long term.

\subsection{Training with driver reactions to traffic signs}\label{belgium}
We begin with a simplified proof-of-concept scenario to establish a baseline. Here, a skilled driver follows familiar routes, occasionally encountering new situations. Based on a sequence of traffic signs, the driver responds by i) stopping, ii) decelerating, or iii) maintaining speed. The continual learning pipeline focuses solely on classifying traffic signs, assuming each has been pre-detected and isolated from the camera feed.

This setup uses a customized version of the Belgium Traffic Signs dataset \cite{mathias2013traffic}, where each image is labeled with one of the three actions. Due to the small dataset size, we apply data augmentation by introducing environmental and equipment distortions such as exposure and saturation issues.

\begin{figure*}[htbp]
\begin{center}
\begin{tabular}{cc}
TRN & ANDE+RAND \\
\includegraphics[width=0.47\linewidth]
{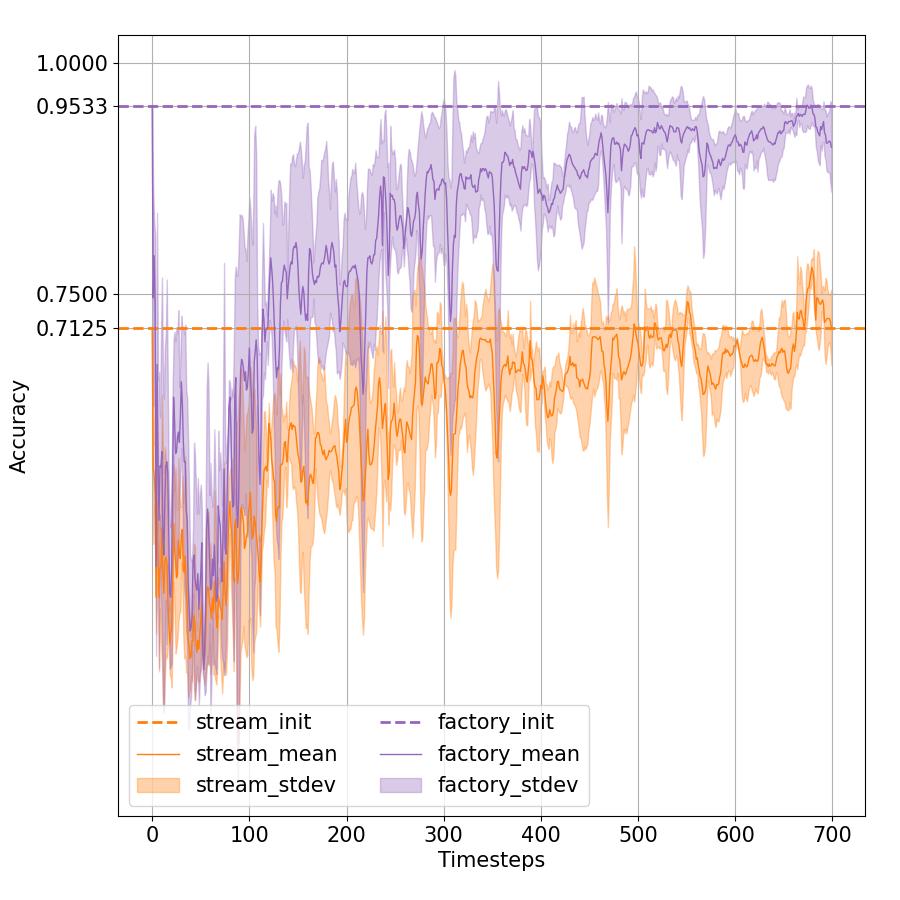}
&
\includegraphics[width=0.47\linewidth]{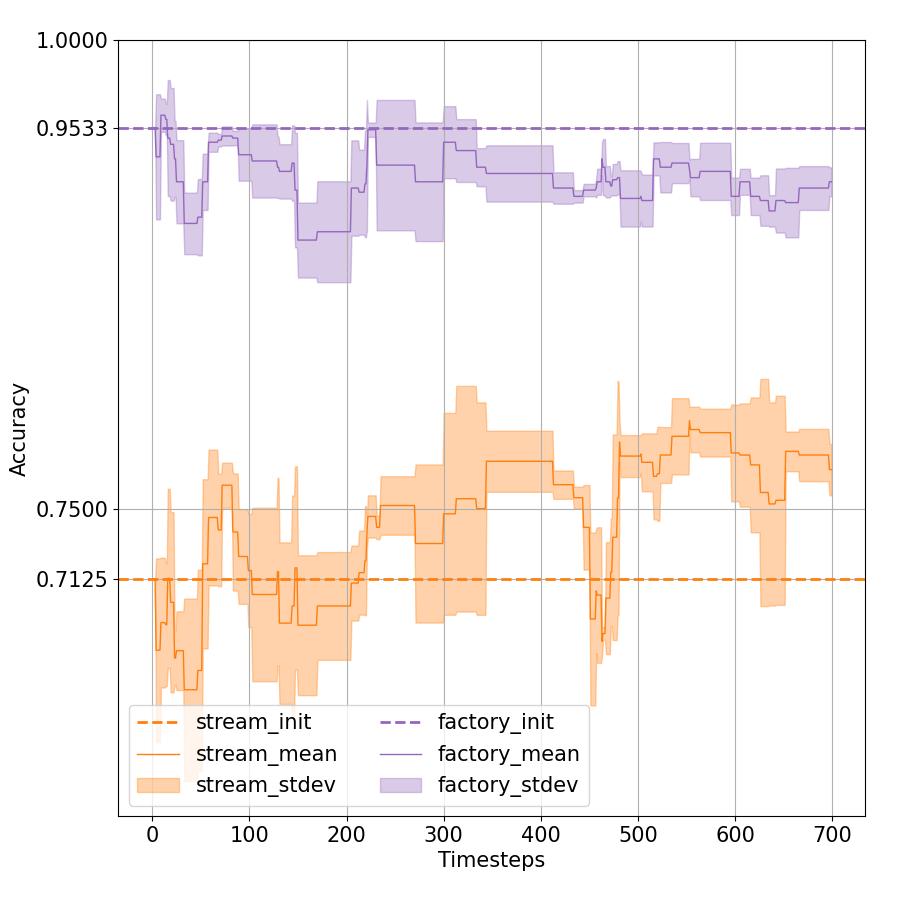} 
\\
ANDE+ECDD & ANDE+ODIN \\
\includegraphics[width=0.47\linewidth]{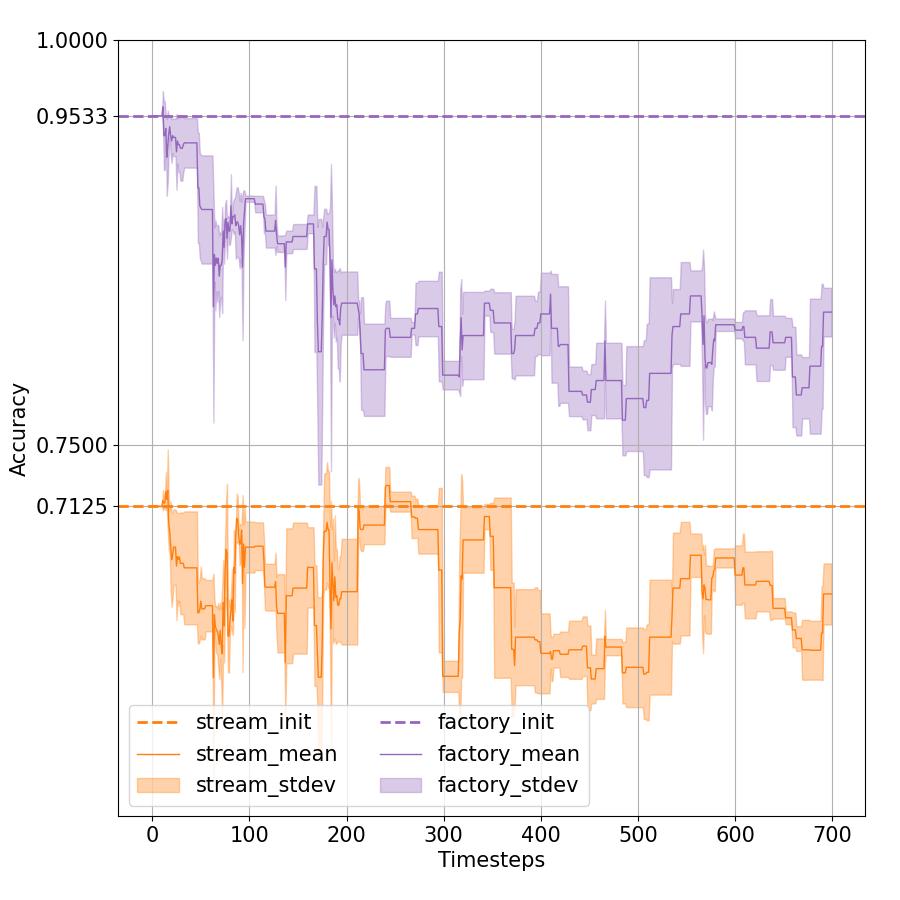}
&
\includegraphics[width=0.47\linewidth]{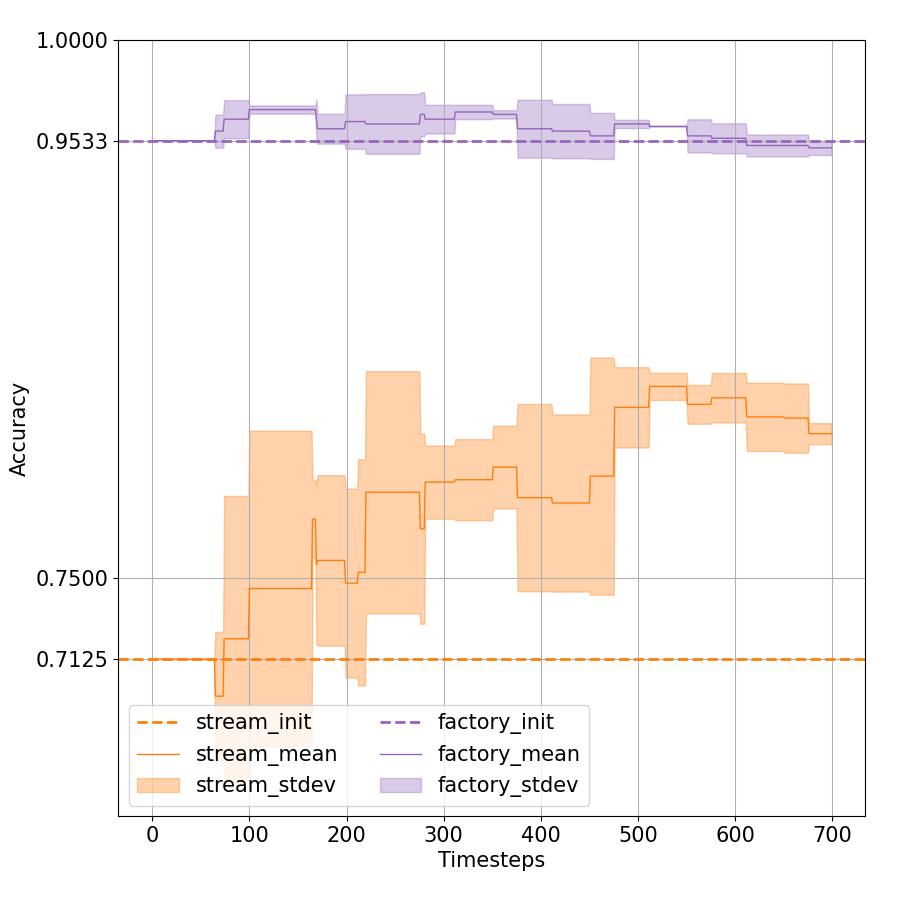}

\\
\end{tabular}
\end{center}
    \caption{Accuracy results for the custom Belgium Traffic Signs dataset with cloud annotation and dataset expansion as described in Table \ref{methods_list} . The dotted lines are the factory and stream accuracy in the first timestep. The results are the mean and standard deviation between three drivers}
    \label{fig:belgium_edge}
\end{figure*}
At each timestep, the model's prediction and the driver’s reaction are recorded simultaneously. The model is pre-trained on approximately 1400 clean, disjoint samples to ensure reasonable initial performance. Training is deliberately limited to prevent overfitting and emphasize the importance of continual updates. The driver is assumed to be 95\% accurate, and all methods are evaluated under this setting.

The stream is a combination of random blocks with different class ratios of [100,50,5] samples per class (between 3 classes) and each sample has 15\% possibility of corruption in the stream. 

The model is plainly composed of two VGG blocks \cite{simonyan2014very} of size 
[32,64] convolutional filters. Each edge device/driver is presented with a stream of 22400 samples with some samples being randomly distorted by using brightness and color intensity, which is a common occurrence when light conditions and camera quality varies and possibly shown more than once. Note that all drivers see a variation of the same stream. The metrics are being calculated with disjoint test data both for factory and stream accuracy. 

Fig. \ref{fig:belgium_edge} shows that omitting rehearsal or anomaly detection significantly reduces accuracy. The following figures present test results for all method combinations during online training in an edge/cloud rehearsal setting. Initial accuracies are $A_{\text{factory}} = 0.9533$ and $A_{\text{stream}} = 0.7125$, reflecting the typical performance gap between factory and deployment. ODIN yields the highest improvement. Table \ref{tab:methods_acc_belgium} lists all combinations with average factory and stream accuracies. Another important improvement is how fast ODIN can increase stream accuracy whilst maintaining both stream and factory accuracy throughout the experiment.

\begin{table}
\caption{Average accuracy on the Belgium Traffic Signs custom dataset with all method combinations. The maximum average is shown in teal.}
\label{tab:methods_acc_belgium}
\centering
\begin{tabular}{c|c|c}
 Strategy   & $\bar{A}_{Factory}$   & $\bar{A}_{Stream}$      \\
\hline
 Initial state       & 0.9533 & 0.7125 \\
 \hline
 TRN        & 0.8124 \textpm 0.0216 & 0.6188 \textpm 0.0148 \\
 \hline
 RH+RAND    & 0.9584 \textpm 0.0008 & 0.7044 \textpm 0.0035 \\
 RH+ODIN    & 0.9447 \textpm 0.0009 & 0.7494 \textpm 0.0146 \\
 RH+ECDD    & 0.9422 \textpm 0.0025 & 0.6952 \textpm 0.0088 \\
 \hline
 ANDE+RAND   & 0.9257 \textpm 0.0048 & 0.7444 \textpm 0.0074 \\
 ANDE+ODIN   & \textbf{\textcolor{teal}{0.9592 \textpm 0.0031}} & \textbf{\textcolor{teal}{0.7856 \textpm 0.0152}} \\
 ANDE+ECDD   & 0.8337 \textpm 0.0024 & 0.659 \textpm 0.0022  \\
\end{tabular}

\end{table}

\subsection{Training for general classification on the road}\label{nuim}
Another important aspect of neural networks within the realms of autonomous or assisted driving is the timely detection and classification of various subjects or objects, without depending on the driver's actions. In this scenario, we assume a system that can work and be updated independently, by using only the model's output response.

\begin{figure*}[hbtp]
\begin{tabular}{cc}
RH+RAND & RH+ODIN \\
\includegraphics[width=0.47\linewidth]{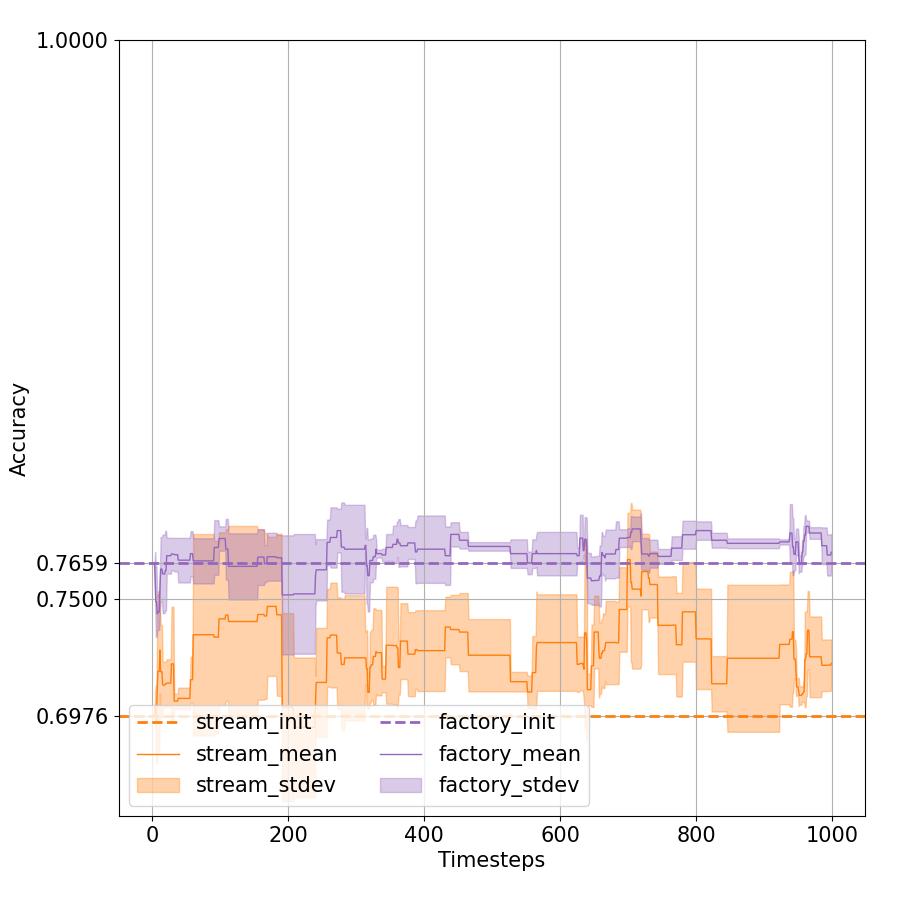}
&
\includegraphics[width=0.47\linewidth]{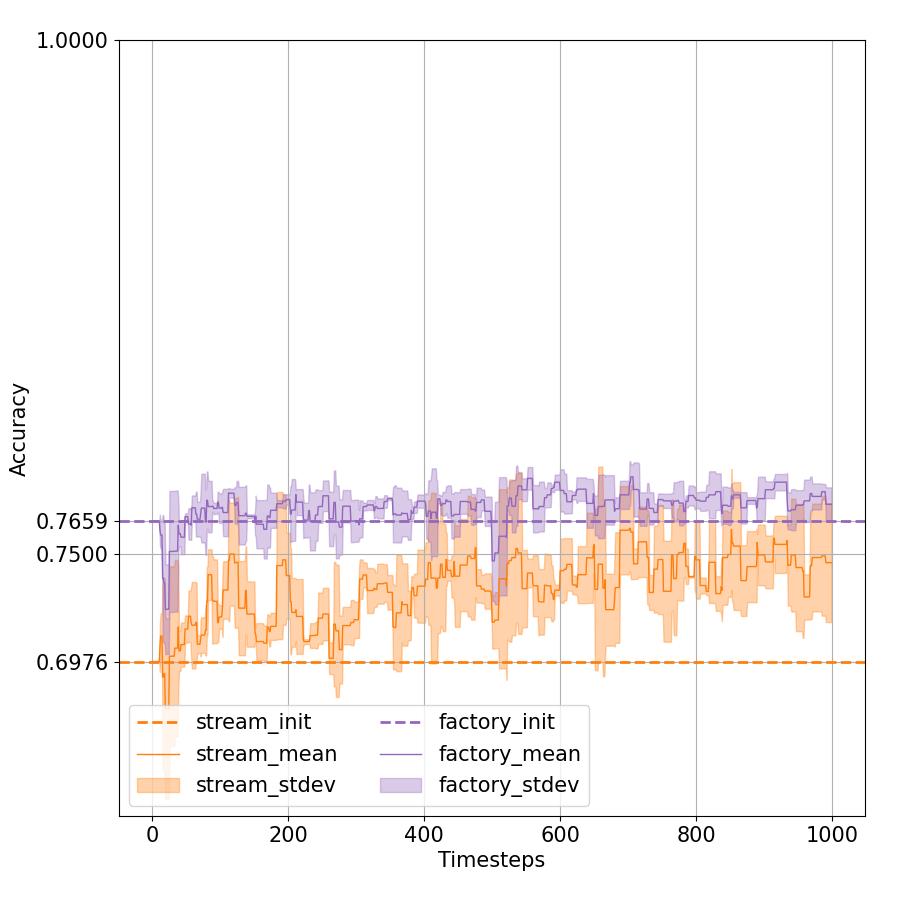} 
\\
ANDE+RAND & ANDE+ODIN \\
\includegraphics[width=0.47\linewidth]{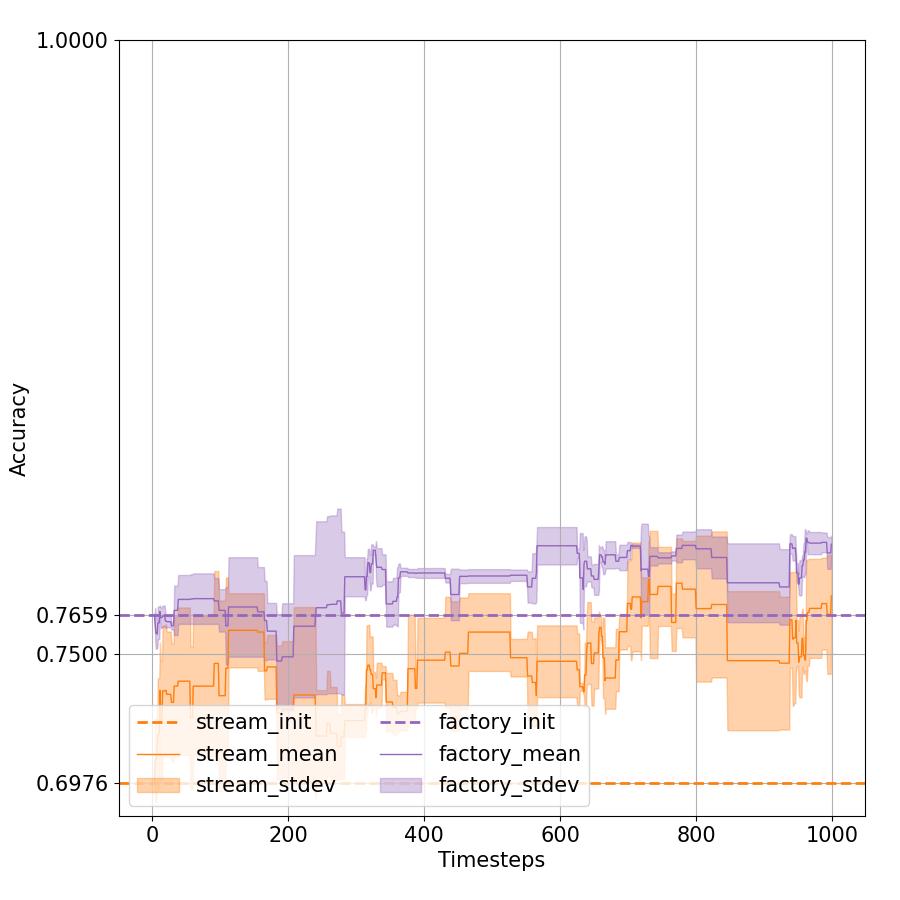}
&
\includegraphics[width=0.47\linewidth]{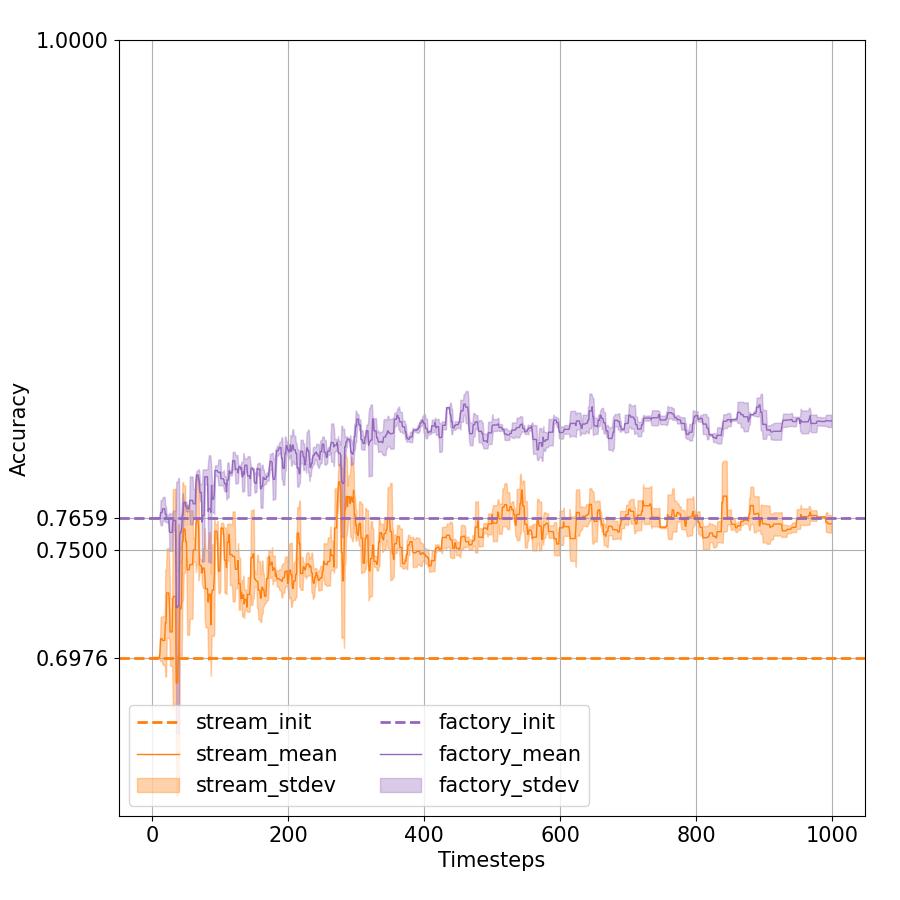}
\end{tabular}
    \caption{Accuracy results for a stream of 1000 timesteps of the nuScenes objects dataset with ODIN and RAND methods as described in Table \ref{methods_list}. The dotted lines are the factory and stream accuracy in the first timestep. The results are the mean and standard deviation between three drivers}
    \label{fig:nuscenes_edge}
\end{figure*}

\begin{table}[hbtp]
    \caption{Number of samples per class for the nuScenes experiment stream}
    \label{tab:nuscenes_dts}
    \centering
    \begin{tabular}{|c|c|c|}
    \hline
    Class name & \#Stream samples & \#Factory samples \\
    \hline
    truck  & 2916 & 589\\
    motorcycle & 1354 & 273\\
    construction & 472 & 95\\
    car &20078 & 4056\\
    bus & 647 & 130\\
    bicycle & 1357 & 274\\
    traffic cone & 6832  & 1380\\
    barrier & 6941 & 1402\\
    construction worker & 1036 & 209\\
    adult & 11998 & 2424\\
    \hline
    total & 53631 & 10832\\
    \hline

    \end{tabular}
\end{table}

The dataset in use is the detected objects from the nuScenes dataset \cite{fong2022panoptic}. We use a similar setup as in the case of Section \ref{belgium}. The stream is composed of frames of various classes which, due to dataset class imbalance, cannot be distributed as in the controlled manner of the stream in \ref{belgium}. Another change is the removal of classes that have a small number of frames, by selecting the top-10 classes in terms of samples in the dataset. In Table \ref{tab:nuscenes_dts} we list the number of pre-training and stream samples per class. In all occasions the samples for training, testing and for factory and stream accuracy calculation are disjoint. In all tests, the combination of cloud annotation, dataset expansion and out-of-distribution detection show a clear increase in classification accuracy both for stream and factory data. 

The results, as shown in Table~\ref{tab:nuscenes_edge} and Figure~\ref{fig:nuscenes_edge} provide a detailed analysis of how each method combination contributes in catastrophic forgetting mitigation with the ANDE+ODIN combination being the most capable both in factory and stream accuracy. ODIN increases and maintains accuracy consistently, while a random sample introduces high deviations in accuracies. Additionally ODIN is able to correct the model in use not only by rising both average accuracies whilst keeping a significantly low deviation as shown in Table~\ref{tab:nuscenes_edge}.

\begin{table}
    \caption{Accuracy results for the nuScenes objects dataset with all available method combinations. The maximum average is shown in teal.}
    \label{tab:nuscenes_edge}
    \centering
\begin{tabular}{c|c|c}
 Strategy   & $\bar{A}_{Factory}$   & $\bar{A}_{Stream}$      \\
 \hline
 Initial state & 0.7659 & 0.6976 \\
  \hline
RH+RAND    & 0.7708 \textpm 0.0018 & 0.7268 \textpm 0.0083 \\
RH+ODIN    & 0.7727 \textpm 0.0006 & 0.7332 \textpm 0.0016 \\
\hline
ANDE+RAND   & 0.7799 \textpm 0.0024 & 0.7478 \textpm 0.0048 \\
ANDE+ODIN   & \textbf{\textcolor{teal}{0.8035 \textpm 0.0006}} & \textbf{\textcolor{teal}{0.7546 \textpm 0.0003}} \\
\end{tabular}
\end{table}

\section{Infection detection}\label{sec:cell}
Considering the scarcity of the appropriate facilities for medical and agricultural assistance in remote areas, detection and classification mechanisms of infections can provide significant assistance in the hands of 
trained individuals. By using an edge device that can isolate images of infected fauna or flora, medical or agricultural professionals can be supported by neural networks for their decisions. 
Note that such a mechanism does not substitute or form the diagnosis or the final verdict of the aforementioned professional. A suggestive mechanism simply provides an important insight for detecting issues for ambiguous samples, assuming that the human input is always prioritized in all occasions. 

Most of the recent works, such as \cite{sharath2020disease} and \cite{asif2023metaheuristics}, despite their significant promise in advancing the field, they do not necessarily evolve dynamically with 
streams of new samples. A continual learning mechanism, can be updated online, providing an evolving platform for all available neural networks. This adaptability can further evolve 
the potential of resource constrained edge devices and eventually provide ubiquitous medical and agricultural assistance across multiple remote locations.

\subsection{Infection detection with cell image classification}\label{malaria}
This scenario considers the challenge of limited medical infrastructure for cell infection detection. Using a continual learning pipeline and an edge device such as a smart microscope, cell images can be diagnosed on-site. While detecting infected cells is a relatively simple binary classification task—where catastrophic forgetting is less prominent—achieving optimal accuracy remains crucial. Such systems can also serve as supplementary diagnostic tools, particularly useful in remote locations.

\begin{figure*}[hbtp]
\begin{center}
\begin{tabular}{cc}
TRN & RH+RAND \\
\includegraphics[width=0.47\linewidth]{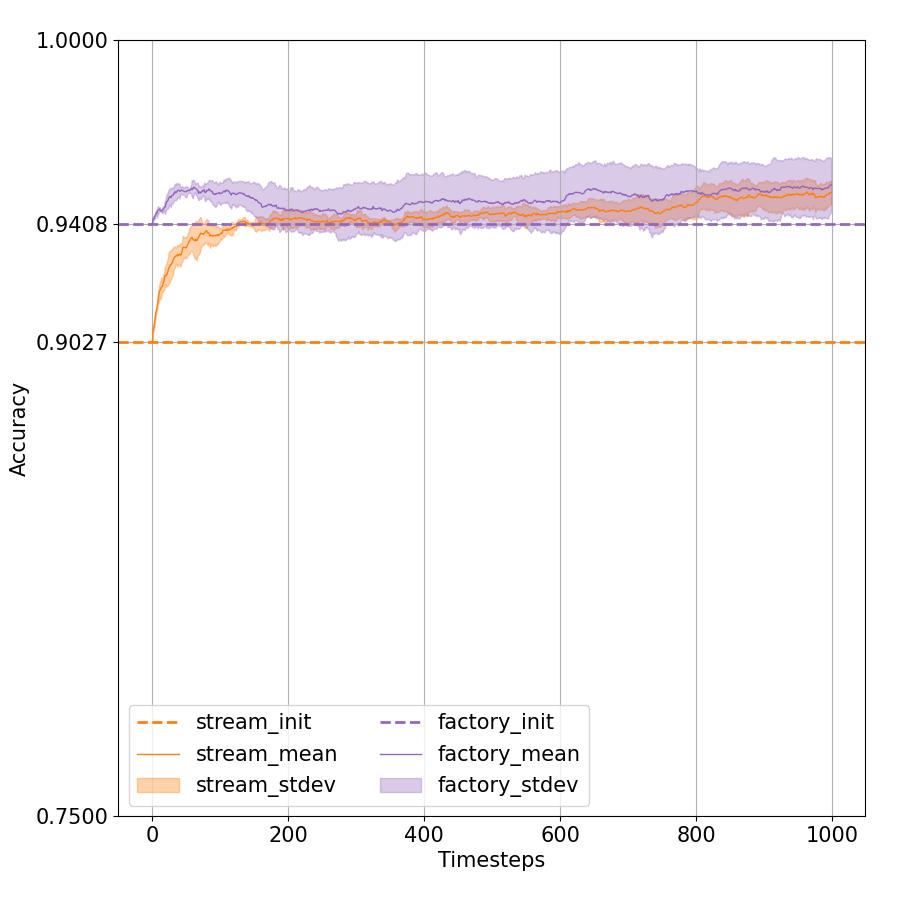}
&
\includegraphics[width=0.47\linewidth]{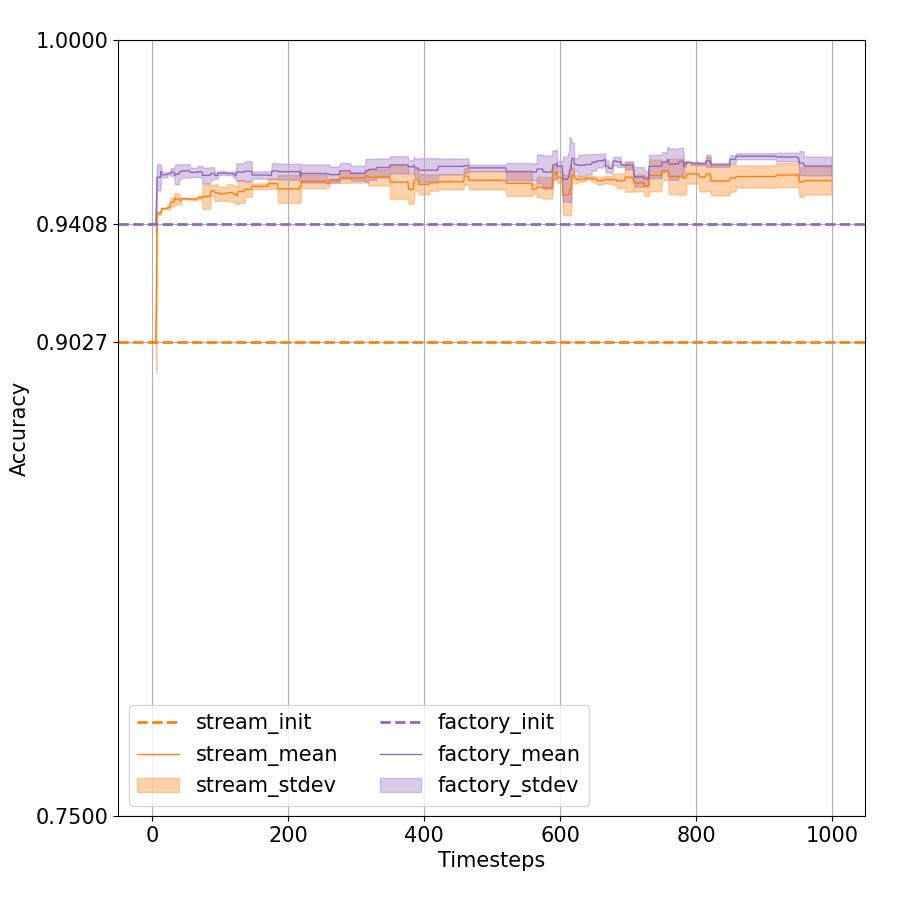} 
\\
RH+ODIN & RH+ECDD \\
\includegraphics[width=0.47\linewidth]{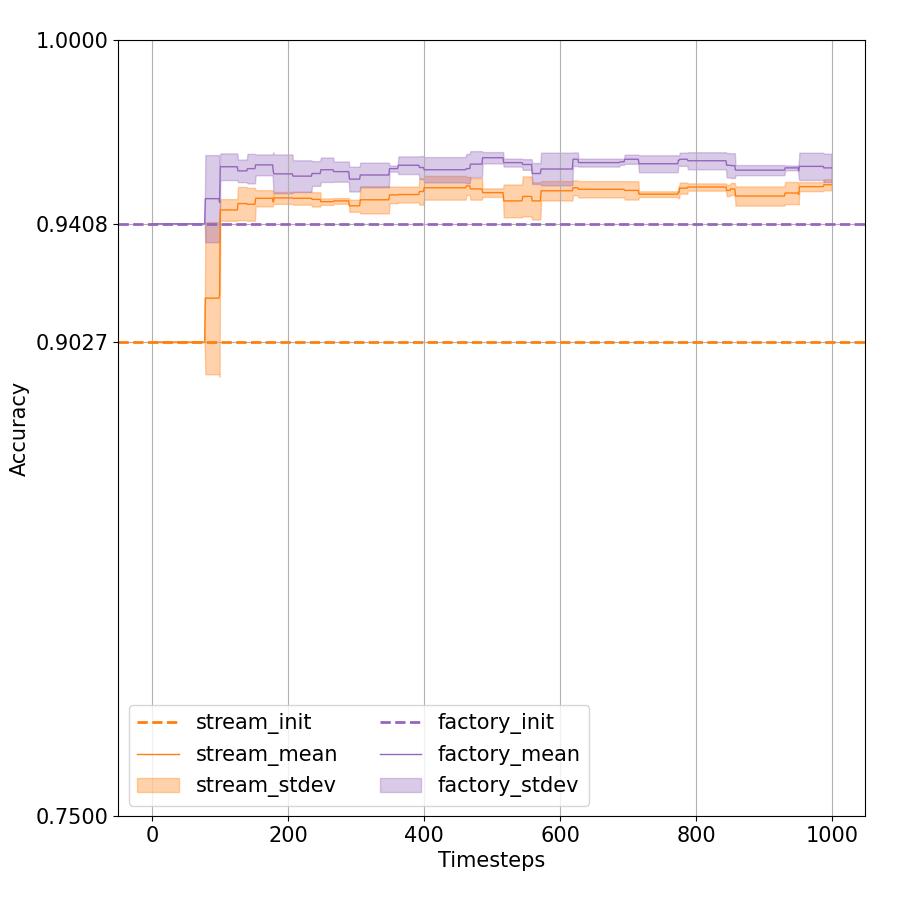}
&
\includegraphics[width=0.47\linewidth]{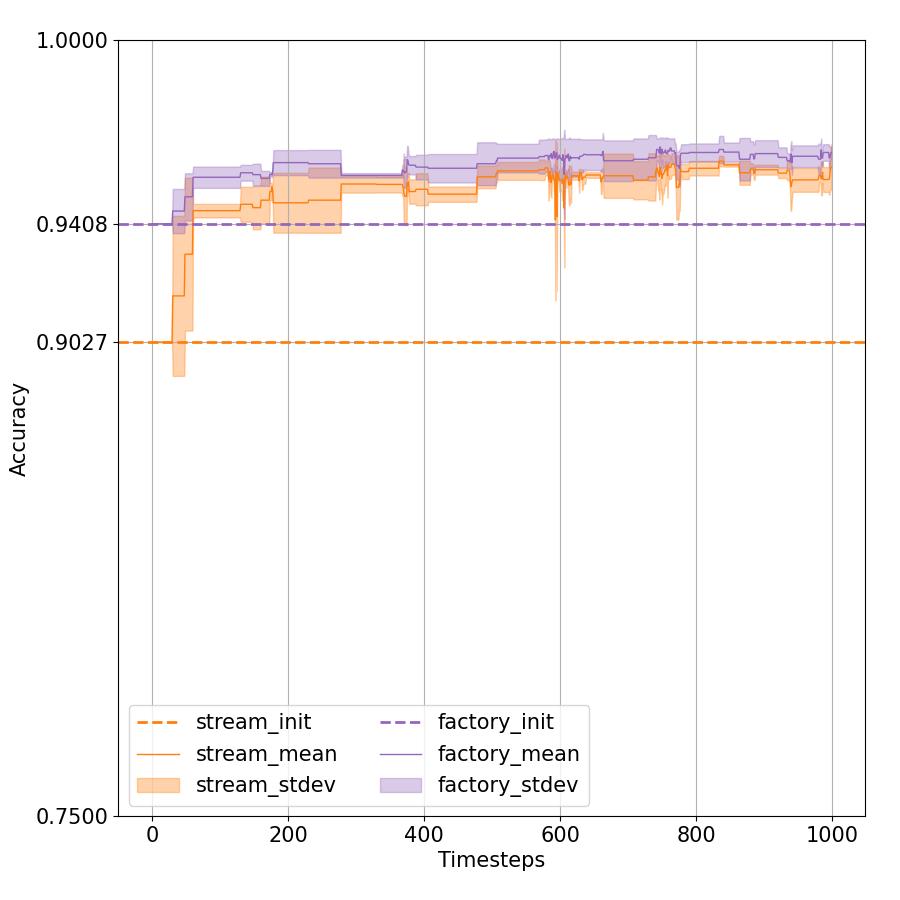}
\end{tabular}
\end{center}
    \caption{Accuracy results for a stream of 1000 timesteps of the Malaria objects dataset with edge-based methods, as described in Table \ref{methods_list}. The dotted lines are the factory and stream accuracy in the first timestep. The results are the mean and standard deviation between three users}
    \label{fig:malaria_edge}
\end{figure*}

In the event of prediction conflicts, medical professionals can provide corrective input to support training. This input, combined with the model output, can trigger a drift detection response. Given the critical nature of medical diagnoses, expert intervention may be necessary when model predictions are uncertain. In such cases, either annotation from a qualified third party or a “user override” mechanism—allowing professionals to label data and retrain the model locally—can be employed.

For our experiment, we use the TensorFlow Malaria dataset \cite{rajaraman2018pre}. To simulate real-world discrepancies, factory data is augmented with slight lighting variations, while stream data includes more severe lighting distortions. As shown in Table~\ref{tab:malaria_edge} and Fig.~\ref{fig:malaria_edge}, catastrophic forgetting is not observed due to the relative simplicity of the task. However, continuous training on every batch would require constant expert input, whereas our methods maintain high accuracy with minimal human intervention. All rehearsal-based approaches perform well, with ECDD and ODIN combinations showing particularly strong performance overall. It is important to note that while edge training does not induce forgetting as strongly as in previous experiments, per-timestep deviation remains a concern. In large-scale disease screening, false positives can have significant consequences, particularly during pandemics. Therefore, minimizing performance variability is also critical.

\begin{table}[hbtp]
    \caption{Accuracy results for the Malaria infected cells dataset with all available method combinations. The maximum average is shown in teal.}
    \label{tab:malaria_edge}
    \centering
\begin{tabular}{c|c|c}
Strategy   & $\bar{A}_{Factory}$   & $\bar{A}_{Stream}$      \\
\hline
Initial state & 0.9408 & 0.9027  \\
\hline
TRN        & 0.949 \textpm 0.0075  & 0.9433 \textpm 0.0021 \\
\hline
RH+RAND    & 0.9588 \textpm 0.0018 & \textbf{\textcolor{teal}{0.9537 \textpm 0.0011}} \\
RH+ODIN    & 0.9574 \textpm 0.0026 & 0.9459 \textpm 0.0005 \\
RH+ECDD    & \textbf{\textcolor{teal}{0.9595 \textpm 0.0029}} & 0.9512 \textpm 0.0017 \\
\end{tabular}
\end{table}

\begin{figure*}[hbtp]
\begin{center}
\begin{tabular}{cc}
TRN & RH+RAND \\
\includegraphics[width=0.47\linewidth]{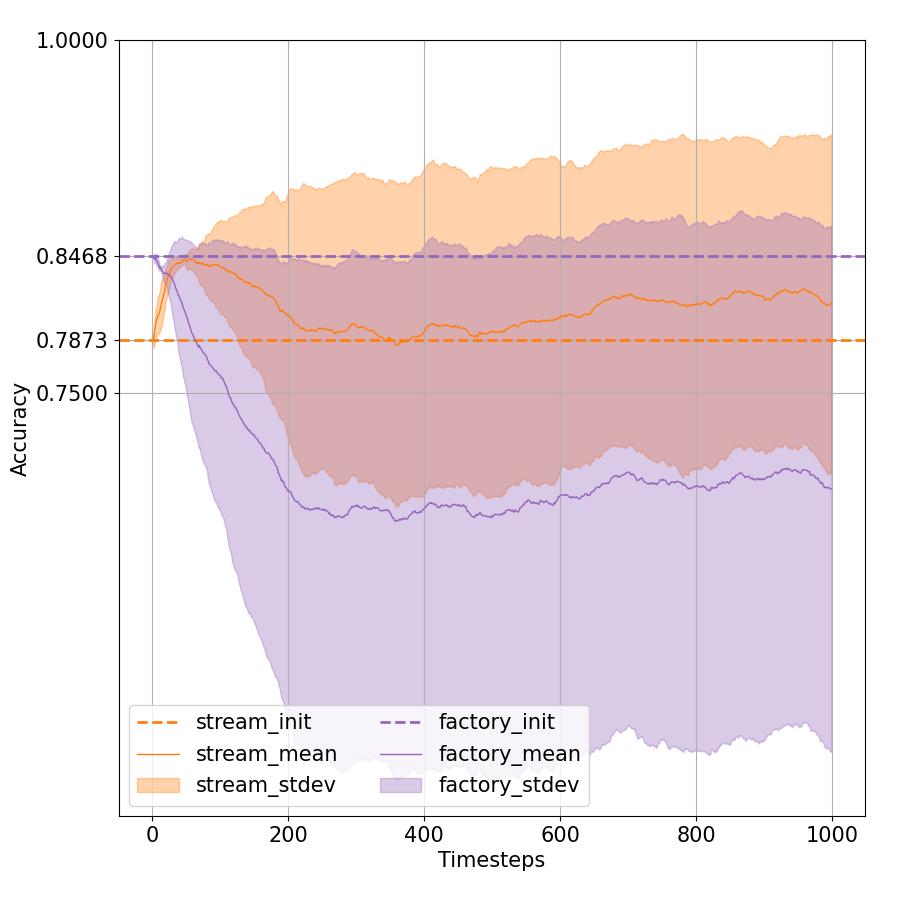}
&
\includegraphics[width=0.47\linewidth]{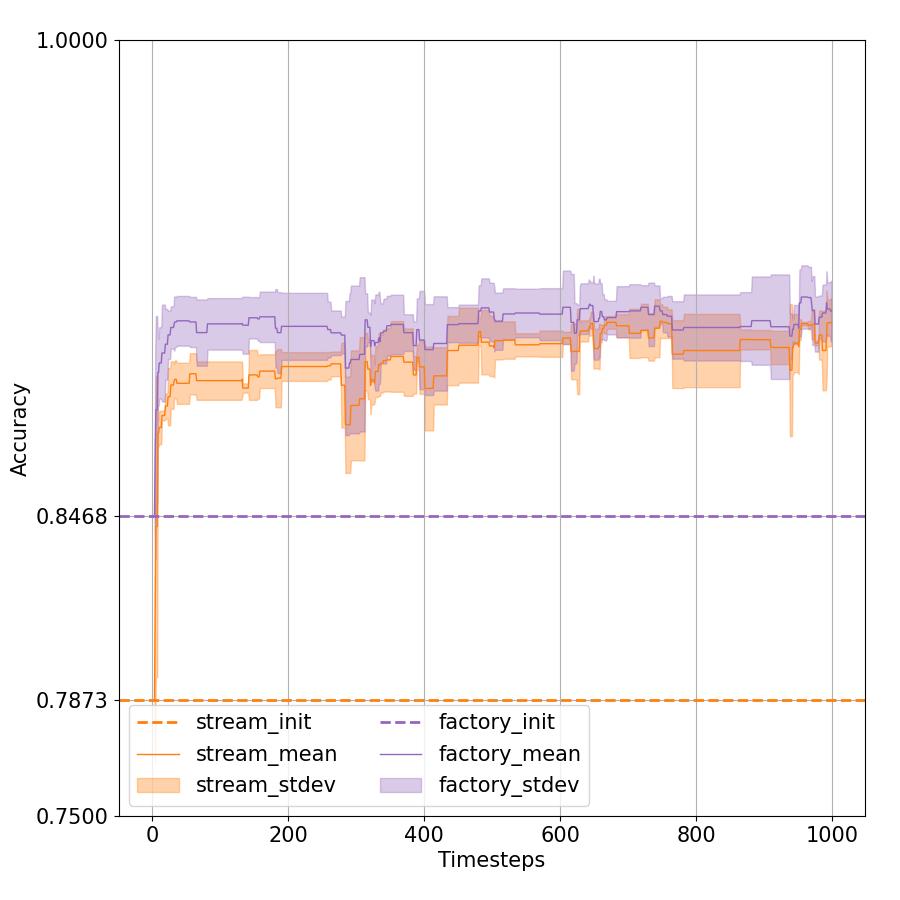} 
\\
RH+ODIN & RH+ECDD \\
\includegraphics[width=0.47\linewidth]{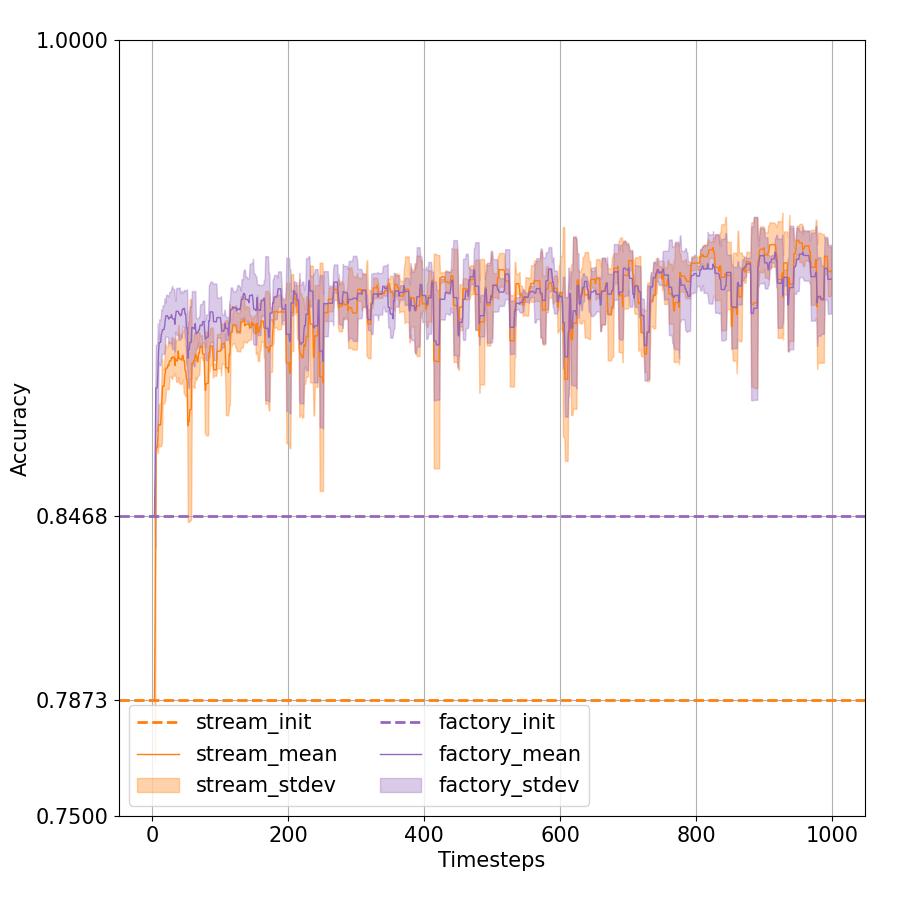}
&
\includegraphics[width=0.47\linewidth]{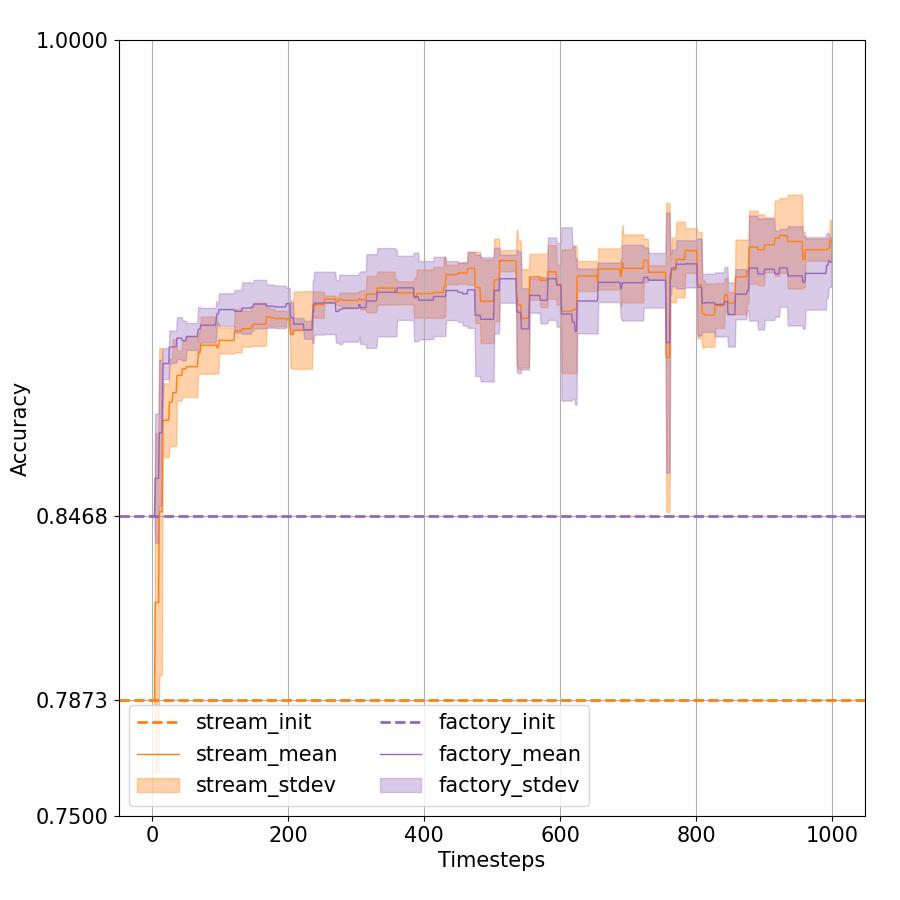}
\end{tabular}
\end{center}
    \caption{Accuracy results for a stream of 1000 timesteps of the Plant Village objects dataset with edge-based methods, as described in Table \ref{methods_list}. The dotted lines are the factory and stream accuracy in the first timestep. The results are the mean and standard deviation between three users}
    \label{fig:plant_village_edge}
\end{figure*}

\subsection{Crop inspection with leaf samples classification}\label{plants}

The objective is the early detection of disease in crops based on drone images of plant foliage. An expert such as a farmer or an agriculturist can use a suggestion machine and provide human feedback when in doubt of a model’s decision. 

When the expert and the model have reached a point of conflict, the questionable samples can be annotated by a third party. We can also use the user annotations as an override, when the expert is adamant about the final verdict. Then the model can train locally if the cloud infrastructure is unavailable.
For the experiment, the TensorFlow Plant village dataset is used \cite{hughes2015open}. We use the similar augmentations as the Malaria example, given that crop inspection samples can be extracted in favorable conditions, unlike in the case of driving. The results show that edge training can cause significant catastrophic forgetting, rendering the model incapable of providing dependable predictions. 

\begin{table}[hbtp]
    \caption{Accuracy results for the Plant Village dataset with all available method combinations. The maximum average is shown in teal.}
    \label{tab:plant_village_edge}
    \centering
\begin{tabular}{c|c|c}
Strategy   & $\bar{A}_{Factory}$   & $\bar{A}_{Stream}$      \\
\hline
Initial state & 0.8468 & 0.7873 \\
\hline
TRN        & 0.6943 \textpm 0.1645 & 0.81 \textpm 0.0947   \\
\hline
RH+RAND    & 0.9082 \textpm 0.0087 & 0.8974 \textpm 0.0031 \\
RH+ODIN    & \textbf{\textcolor{teal}{0.9168 \textpm 0.0059}} & 0.9145 \textpm 0.0025 \\
RH+ECDD    & 0.916 \textpm 0.009   & \textbf{\textcolor{teal}{0.9168 \textpm 0.0027}} \\
\end{tabular}
\end{table}

The results in Figure~\ref{fig:plant_village_edge} and Table~\ref{tab:plant_village_edge} 
show how catastrophic forgetting has been induced when using the TRN baseline, while the RH+ODIN and RH+ECDD methods can adequately mitigate the effect. TRN shows significant amount of deviations and a considerable drop in average accuracy. This phenomenon is considerably mitigated by ODIN and ECDD with a considerable increase in both stream and factory accuracies on average

\FloatBarrier
\section{Discussion}\label{sec:disc}
This study presented a framework for applying rehearsal-based continual learning methods in cloud-edge scenarios for assisting professionals in a variety of real life application such as assisted driving and supplementary infection detection.
Our experiments demonstrate that selective sampling and rehearsal-based model updates can support adaptation under distribution shifts while mitigating catastrophic forgetting in the evaluated datasets. The proposed edge-cloud pipeline provides an architectural setting for incorporating these mechanisms into IoT applications.
According to our experiments, a combination of rehearsal methods, drift detection or out-of-distribution detection can be very potent, not only for catastrophic forgetting mitigation, but also for evolving 
the classification ability of a pre-trained model. It also important to note that there is not a specific approach that can adapt to all possible scenarios. The application each method has its own benefits and limitations. Of course some limitations can arise such as storing samples that require user permission. Additionally these methods can possibly be further improved given that both drift detection and OoD literature shows promise and growth

\bibliographystyle{IEEEtran}
\bibliography{bibliography}

\end{document}